\documentclass[]{fairmeta}
\usepackage[T1]{fontenc}

\usepackage{graphicx}
\usepackage{amsmath,amssymb}
\usepackage{booktabs}
\usepackage{multirow}
\usepackage{tcolorbox}
\usepackage{algorithm}
\usepackage{algorithmic}
\usepackage{xcolor}
\usepackage{tikz}
\usepackage{pgfplots}
\usepackage{placeins}
\usepackage{natbib}

\pgfplotsset{compat=newest}
\usetikzlibrary{positioning,arrows.meta,calc}
\tikzset{
  bank/.style={circle, draw, thick, minimum size=0.9cm, font=\footnotesize\sffamily},
  edgea/.style={->, line width=0.9pt, color=red!70},
  edgeb/.style={->, line width=0.9pt, color=orange!70},
  edgec/.style={->, line width=0.7pt, color=gray!60, dashed},
  lbl/.style={font=\scriptsize\sffamily, inner sep=1pt}
}

\newcommand{\best}[1]{{\bfseries\boldmath #1}}
\newcommand{\secondbest}[1]{\underline{#1}}
\newcommand{\pmv}[2]{\(\,#1 \pm #2\,\)}

\title{CausalGraphX: A Counterfactual Graph Neural Network Framework for Explainable Systemic Risk Assessment}

\author[1\dagger,3]{Rabimba Karanjai}
\author[3]{Hemanth Hegadehalli Madhavarao}
\author[2]{Lei Xu}
\author[1]{Weidong Shi}

\affiliation[1]{University of Houston}
\affiliation[2]{Kent State University}
\affiliation[3]{PayPal Inc.}
\contribution[\dagger]{rkaranjai@paypal.com}

\abstract{
The interconnected nature of global financial systems makes them vulnerable to systemic risks, where the failure of a few institutions can trigger catastrophic cascading defaults. Traditional risk models often fail to capture the complex, non-linear dynamics of these networks. While Graph Neural Networks (GNNs) have shown promise in modeling relational data, they primarily learn correlative patterns and function as black boxes, offering little insight into the causal mechanisms of shock propagation. This limitation is critical for regulators who require explainable models to perform stress tests and devise effective interventions. We introduce CausalGraphX, a novel framework that integrates GNNs with counterfactual reasoning to provide explainable assessments of systemic risk. CausalGraphX employs a Graph Attention mechanism to learn representations of institutional vulnerability and uses an adversarial regularization technique to ensure these representations capture causal drivers rather than spurious correlations. Furthermore, we propose an optimization-based approach to generate counterfactual explanations, answering questions such as, "What minimum capital injection would have prevented Bank A's default under a specific stress scenario?" We validate CausalGraphX on large-scale synthetic financial networks. Our results demonstrate that CausalGraphX significantly outperforms traditional and deep learning baselines in predicting cascading defaults while providing sparse, plausible, and actionable counterfactual explanations.
}

\date{\today}

\begin{document}
\maketitle

\section{Introduction}

The global financial system is characterized by a dense web of interconnections, primarily through interbank lending, derivatives, and asset holdings. While this interconnectedness promotes market efficiency, it also exposes the system to significant systemic risk—the risk that the failure of one or a few institutions can propagate rapidly, leading to a system-wide crisis \citep{acemoglu2015systemic}. The 2008 financial crisis underscored the inadequacy of traditional risk assessment models in anticipating such events.

In response, the research community developed network-based models to quantify systemic risk. Foundational models, such as the seminal work by \citet{eisenberg2001systemic}, provided insights into default cascades under simplified assumptions. Subsequent approaches, like DebtRank \citep{battiston2012debtrank}, offered more sophisticated metrics for institutional importance. However, these methods often rely on strong linearity assumptions and struggle to leverage the rich feature sets available for financial institutions.

The advent of deep learning, particularly Graph Neural Networks (GNNs), has opened new avenues for modeling complex relational systems. GNNs can effectively learn node representations by aggregating information from their neighbors, making them naturally suited for analyzing financial networks \citep{bargigli2021network}.

Despite their predictive power, standard GNNs present two critical limitations in the context of systemic risk regulation:

\begin{enumerate}
    \item \textbf{Lack of Explainability:} GNNs typically operate as black boxes. While they might predict a default, they cannot easily articulate \textit{why}, or which specific exposures are the primary drivers of risk. Regulators require transparent models to design effective policy interventions.
    \item \textbf{Correlative vs. Causal Inference:} Standard GNNs learn correlations. In financial networks, this can lead to spurious correlations being mistaken for causal drivers of distress. For effective stress testing ("what-if" scenarios), models must capture the underlying causal mechanisms of shock propagation.
\end{enumerate}

To address these limitations, we propose \textbf{CausalGraphX}, a novel Counterfactual Graph Neural Network framework for explainable systemic risk assessment. CausalGraphX is designed not only to predict the likelihood of cascading defaults but also to provide intuitive, actionable explanations grounded in causality.

Our framework consists of two main components. The first is a Causal Representation Learner based on a Graph Attention Network (GAT) architecture \citep{velickovic2018graph}. We augment the standard GAT training objective with an adversarial regularization term inspired by advances in causal representation learning \citep{johansson2016learning}. This encourages the model to learn embeddings that are predictive of default while remaining invariant to confounding factors.

The second component is a Counterfactual Explanation Generator. This module generates counterfactuals—hypothetical scenarios where a specific outcome (a bank's default) is reversed by a minimal change in the input (an increase in capital reserves). This allows risk managers to identify the most efficient interventions to mitigate systemic risk.

Our main contributions are summarized as follows:

\begin{itemize}
    \item We introduce CausalGraphX, a framework unifying GNN-based systemic risk modeling with counterfactual explainability.
    \item We propose an adversarial regularization approach to enhance the causal validity of learned graph representations in financial networks.
    \item We develop an efficient optimization algorithm for generating sparse and plausible counterfactual explanations for systemic risk events, providing actionable insights for regulators.
    \item We empirically demonstrate that CausalGraphX achieves superior predictive performance compared to state-of-the-art baselines while providing high-quality explanations.
\end{itemize}

\section{Related Work}

Our work draws upon three main streams of prior work: systemic risk modeling, GNNs in finance, and explainable AI (XAI) with a focus on causality.

\subsection{Systemic Risk Modeling}

The modeling of systemic risk has evolved significantly. The foundational model by \citet{eisenberg2001systemic} (E-N) established a clearing mechanism for interbank liabilities. DebtRank \citep{battiston2012debtrank} introduced a feedback mechanism to measure the systemic importance of institutions. \citet{adrian2016covar} proposed CoVaR (Conditional Value at Risk) to measure the contribution of an institution to the overall system risk when that institution is in distress. While insightful, these methods often struggle to incorporate high-dimensional node features and are primarily diagnostic rather than predictive.

\subsection{Graph Neural Networks in Finance}

The application of GNNs in finance is rapidly growing, applied to tasks such as anti-money laundering (AML) \citep{weber2019anti} and stock market prediction \citep{feng2019temporal}. In the context of systemic risk, recent works have started exploring GNNs. \citet{bargigli2021network} utilized Graph Convolutional Networks (GCNs) to predict bank defaults based on interbank exposures. However, these applications generally focus purely on prediction accuracy and do not address the critical need for explainability or causal inference.

\subsection{Explainable AI and Causality}

Explainable AI (XAI) methods aim to interpret complex models. Post-hoc methods like LIME \citep{ribeiro2016should} and SHAP \citep{lundberg2017unified} provide local explanations but can be unstable and often highlight correlations rather than causal drivers \citep{rudin2019stop}.

Counterfactual explanations \citep{wachter2017counterfactual} offer a more intuitive approach, identifying the smallest change to the input features required to change the model's prediction. This aligns well with the concept of interventions in causal inference \citep{pearl2009causality}. Recently, methods like CF-GNNExplainer \citep{lucic2022cf} aim to identify influential subgraphs for GNN predictions. Our work differs by focusing on modifying node features (capital levels) and edge weights (exposures) to prevent systemic events, and we explicitly incorporate causal regularization during training.

\section{Methodology: The CausalGraphX Framework}

We introduce the architecture of CausalGraphX, comprising two stages: (1) Causal Representation Learning, and (2) Counterfactual Explanation Generation.

\begin{figure}
    \centering
    \includegraphics[width=1\linewidth]{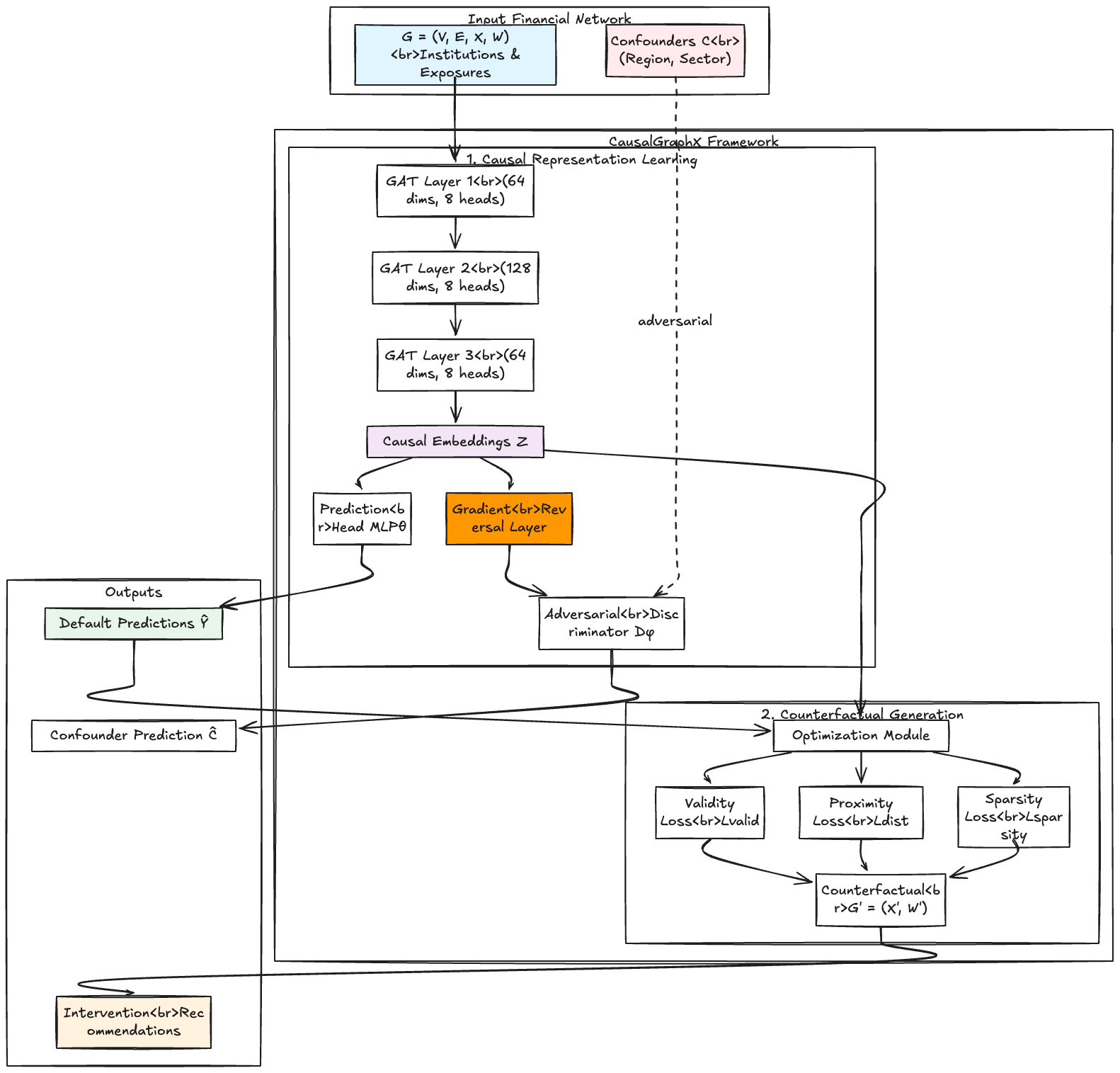}
    \caption{Architecture of CausalGraphX}
    \label{fig:CausalGraphX}
\end{figure}

Figure \ref{fig:CausalGraphX} shows the overall architecture of the framework.

\subsection{Problem Formulation}

We model the financial system as a directed graph $G = (\mathcal{V}, \mathcal{E}, X, W)$, where $\mathcal{V}$ is the set of $N$ institutions (nodes), and $\mathcal{E}$ is the set of connections (edges). $X \in \mathbb{R}^{N \times F}$ is the feature matrix (capital reserves, leverage). $W \in \mathbb{R}^{N \times N}$ is the weighted adjacency matrix, where $W_{ij}$ represents the exposure of institution $i$ to institution $j$. Given an initial shock scenario, the goal is to predict the final state of each institution $Y \in \{0, 1\}^N$, where $Y_i=1$ indicates default.

\subsection{Causal Representation Learning}

The core of CausalGraphX is a GNN model $f_\theta(G)$ parameterized by $\theta$, which learns latent representations $Z \in \mathbb{R}^{N \times D}$. We utilize a Graph Attention Network (GAT) architecture due to its ability to assign different importance to different neighbors, crucial for modeling heterogeneous financial exposures.

\subsubsection{Graph Attention Mechanism}

The GAT layer updates the representation of node $i$ by attending over its neighbors $\mathcal{N}(i)$. The attention coefficient $\alpha_{ij}$ between node $i$ and neighbor $j$ is calculated, incorporating the edge weight $W_{ij}$:

\begin{equation}
e_{ij} = \text{LeakyReLU}(\vec{a}^T [H_i || H_j || W_{ij}])
\end{equation}
\begin{equation}
\alpha_{ij} = \text{softmax}_j(e_{ij}) = \frac{\exp(e_{ij})}{\sum_{k \in \mathcal{N}(i)} \exp(e_{ik})}
\end{equation}

where $H$ are the input representations (initially $X$), $||$ denotes concatenation, and $\vec{a}$ is a learnable weight vector. The updated representation $H'_i$ is:

\begin{equation}
H'_i = \sigma\left(\sum_{j \in \mathcal{N}(i)} \alpha_{ij} \mathbf{W} H_j\right)
\end{equation}

After $L$ layers, we obtain the final embeddings $Z$. A prediction head (MLP) predicts the probability of default: $\hat{Y} = \text{MLP}(Z)$.

\subsubsection{Adversarial Causal Regularization}

Standard GNN training minimizes the prediction loss, typically binary cross-entropy (BCE):

\begin{equation}
\mathcal{L}_{pred} = \sum_{i=1}^N \text{BCE}(Y_i, \hat{Y}_i)
\end{equation}

This objective encourages exploiting all correlations, including spurious ones. To ensure $Z$ captures causal drivers and is robust to interventions, we introduce adversarial regularization. We aim for invariance of $Z$ with respect to confounding variables $C$.

We introduce a discriminator (adversary) $D_\phi$, parameterized by $\phi$, which attempts to predict the confounder $C$ from $Z$. The GNN $f_\theta$ tries to learn embeddings $Z$ that are predictive of $Y$ but minimally predictive of $C$. This is formulated as a minimax game:

\begin{equation}
\min_\theta \max_\phi \mathcal{L}_{pred}(\theta) - \lambda \mathcal{L}_{adv}(\theta, \phi)
\end{equation}

where $\mathcal{L}_{adv}$ is the loss of the adversary (cross-entropy for predicting $C$), and $\lambda$ balances the trade-off. By enforcing this invariance, the model is forced to rely on causal pathways (direct exposures) rather than spurious correlations related to $C$.

\subsection{Counterfactual Explanation Generator}

Once $f_\theta$ is trained, we generate counterfactual explanations. A counterfactual addresses: "Given a predicted default ($Y_A=1$), what is the minimum change to $X$ or $W$ such that the model predicts survival ($Y'_A=0$)?"

We formulate the generation of a counterfactual instance $G' = (\mathcal{V}, \mathcal{E}, X', W')$ as an optimization problem:

\begin{equation}
\begin{aligned}
G' = \arg\min_{X', W'} \quad & \mathcal{L}_{valid}(f_\theta(G'), Y_{target}) + \\
& \gamma_1 \cdot \mathcal{L}_{dist}(G, G') + \\
& \gamma_2 \cdot \mathcal{L}_{sparsity}(G')
\end{aligned}
\label{eq:counterfactual_opt}
\end{equation}

The objective consists of three terms:

\subsubsection{Validity Loss ($\mathcal{L}_{valid}$)}
Ensures $G'$ results in the target outcome $Y_{target}$. We use the hinge loss:

\begin{equation}
\mathcal{L}_{valid} = \max(0, f_\theta(G')_i - Y_{target} + \kappa)
\end{equation}
where $\kappa$ is a margin parameter.

\subsubsection{Distance Loss ($\mathcal{L}_{dist}$)}
Ensures proximity (the counterfactual $G'$ should be close to $G$). We use the L2 norm for feature/edge changes, often weighted (by inverse Median Absolute Deviation, MAD) to account for feature scales.

\begin{equation}
\mathcal{L}_{dist} = ||X - X'||_{MAD\_L2} + ||W - W'||_{MAD\_L2}
\end{equation}

\subsubsection{Sparsity Loss ($\mathcal{L}_{sparsity}$)}
Encourages changes to only a few features or edges (interpretable intervention). We use the L1 norm:

\begin{equation}
\mathcal{L}_{sparsity} = ||X - X'||_1 + ||W - W'||_1
\end{equation}

\subsubsection{Optimization Procedure}

The optimization problem in Equation \ref{eq:counterfactual_opt} is solved using a gradient-based approach (Adam). Since the GNN is differentiable, we can backpropagate gradients from the loss to the input features and weights. The procedure is summarized in Algorithm 1.

\begin{algorithm}[tb]
\caption{Counterfactual Generation via Optimization}
\label{alg:counterfactual}
\textbf{Input}: Trained model $f_\theta$, Original Graph $G=(X, W)$, Target $Y_{target}$, Hyperparameters $\gamma_1, \gamma_2, \kappa$
\textbf{Output}: Counterfactual Graph $G'=(X', W')$
\begin{algorithmic}[1]
\STATE Initialize $X' \leftarrow X$, $W' \leftarrow W$
\STATE Initialize optimizer (Adam)
\FOR{$t=1$ to $T_{max}$}
    \STATE Calculate prediction $\hat{Y}' = f_\theta(G')$
    \STATE Calculate $\mathcal{L}_{valid}, \mathcal{L}_{dist}, \mathcal{L}_{sparsity}$
    \STATE Calculate total loss $\mathcal{L}_{total}$ (Eq. \ref{eq:counterfactual_opt})
    \IF{$\mathcal{L}_{valid} = 0$}
        \STATE BREAK
    \ENDIF
    \STATE Compute gradients $\nabla_{X'} \mathcal{L}_{total}$ and $\nabla_{W'} \mathcal{L}_{total}$
    \STATE Update $X', W'$ using the optimizer
\ENDFOR
\STATE \textbf{return} $G'$
\end{algorithmic}
\end{algorithm}

\section{Experimental Setup}

Evaluating systemic risk models is challenging due to data scarcity. Following established practices \citep{bargigli2021network}, we rely on synthetic datasets generated using realistic network formation models and financial simulators.

\subsection{Datasets}

The generation process involves network formation and shock simulation.

\subsubsection{Network Formation}

We simulate interbank networks using two models to ensure robustness:

\begin{itemize}
    \item \textbf{Erd\H{o}s–R\'{e}nyi (ER) Model:} Represents a random network structure.
    \item \textbf{Barab\'{a}si–Albert (BA) Model:} Generates scale-free networks with preferential attachment, leading to "hubs" (highly connected institutions), mirroring real-world core-periphery structures.
\end{itemize}

We generate networks with $N=1000$ institutions. Node features $X$ are drawn from realistic distributions. Edge weights $W$ (exposures) follow a log-normal distribution.

\subsubsection{Shock Simulation and Ground Truth}

To generate ground truth labels $Y$, we use a modified Eisenberg-Noe simulator \citep{eisenberg2001systemic}. We introduce an initial shock by defaulting a random subset of institutions (1\%). The simulator propagates distress until a new equilibrium is reached.

To test causal regularization, we explicitly introduce a confounding variable $C$ (a regional identifier). We generate $X$ and $Y$ such that they are both influenced by $C$, creating spurious correlations that do not reflect the E-N simulator's causal mechanism.

\subsection{Baselines}

We compare CausalGraphX against:

\begin{itemize}
    \item \textbf{Logistic Regression (LR):} Uses node features $X$ and basic network statistics.
    \item \textbf{DebtRank (DR):} A traditional systemic risk measure \citep{battiston2012debtrank}.
    \item \textbf{Graph Convolutional Network (GCN):} The standard GCN model \citep{kipf2017semi}.
    \item \textbf{Graph Attention Network (GAT):} Identical architecture to CausalGraphX but without causal regularization ($\lambda=0$).
\end{itemize}

\subsection{Evaluation Metrics}

We use the Area Under the ROC curve (AUC) and F1-score to measure the ability to predict defaults.

We evaluate counterfactual using:
\begin{itemize}
    \item \textbf{Validity:} Percentage of generated counterfactuals $G'$ that successfully change the prediction.
    \item \textbf{Proximity:} Average distance between $G$ and $G'$ (lower is better).
    \item \textbf{Sparsity:} Average number of features/exposures changed (lower is better).
    \item \textbf{Plausibility (Causal Correctness):} On the synthetic dataset where the ground truth causal mechanism is known, we measure how often the proposed intervention aligns with the true causal drivers of the default.
\end{itemize}

\section{Results and Discussion}

We present a comprehensive empirical evaluation of CausalGraphX on both synthetic and semi-synthetic financial network datasets. Our experiments were conducted on an NVIDIA A100 GPU with 40GB memory, using PyTorch 1.13 and PyTorch Geometric 2.2. All experiments are averaged over 5 random seeds with different initial shock scenarios.

\subsection{Experimental Setup Details}

We generate 10,000 synthetic financial networks for each topology (ER and BA) with $N \in \{500, 1000, 2000\}$ institutions. Node features include 12 financial indicators: Tier 1 capital ratio, leverage ratio, liquidity coverage ratio, non-performing loan ratio, return on assets, cost-to-income ratio, loan-to-deposit ratio, net interest margin, total assets (log-scaled), derivatives exposure, foreign currency exposure, and interbank lending ratio. Edge weights follow a log-normal distribution $\mathcal{LN}(\mu=10^8, \sigma=2)$ representing exposure amounts in USD.

CausalGraphX uses 3 GAT layers with hidden dimensions, 8 attention heads in each layer, dropout rate of 0.3, and Leaky ReLU activation ($\alpha=0.2$). The adversarial discriminator is a 3-layer MLP where |C| is the number of confounding categories.

\subsection{Predictive Performance}

Table \ref{tab:predictive_performance} presents the comprehensive evaluation across different network sizes and topologies.

\begin{table*}[t]
\centering
\caption{Predictive Performance for Systemic Risk Assessment. Results averaged over 5 seeds with standard deviations. Bold indicates best performance, underline indicates second best.}
\label{tab:predictive_performance}
\setlength{\tabcolsep}{5pt}     
\renewcommand{\arraystretch}{1.15} 

\footnotesize
\resizebox{\textwidth}{!}{%
\begin{tabular}{@{}lcccccccc@{}}
\toprule
\multirow{3}{*}{\textbf{Model}} &
\multicolumn{4}{c}{\textbf{Erd\H{o}s–R\'{e}nyi (ER) Network}} &
\multicolumn{4}{c}{\textbf{Barab\'{a}si–Albert (BA) Network}} \\
\cmidrule(lr){2-5} \cmidrule(lr){6-9}
& \multicolumn{2}{c}{\(N{=}1000\)} & \multicolumn{2}{c}{\(N{=}2000\)} &
  \multicolumn{2}{c}{\(N{=}1000\)} & \multicolumn{2}{c}{\(N{=}2000\)} \\
\cmidrule(lr){2-3} \cmidrule(lr){4-5} \cmidrule(lr){6-7} \cmidrule(lr){8-9}
& \textbf{AUC} & \textbf{F1} & \textbf{AUC} & \textbf{F1} & \textbf{AUC} & \textbf{F1} & \textbf{AUC} & \textbf{F1} \\
\midrule
Logistic Regression & \pmv{0.724}{0.031} & \pmv{0.652}{0.042} & \pmv{0.738}{0.028} & \pmv{0.671}{0.038} & \pmv{0.706}{0.044} & \pmv{0.634}{0.051} & \pmv{0.719}{0.039} & \pmv{0.648}{0.047} \\
DebtRank            & \pmv{0.761}{0.023} & \pmv{0.698}{0.031} & \pmv{0.778}{0.021} & \pmv{0.715}{0.029} & \pmv{0.752}{0.032} & \pmv{0.683}{0.041} & \pmv{0.769}{0.028} & \pmv{0.701}{0.036} \\
GCN                 & \pmv{0.843}{0.018} & \pmv{0.782}{0.024} & \pmv{0.856}{0.016} & \pmv{0.798}{0.022} & \pmv{0.821}{0.027} & \pmv{0.764}{0.034} & \pmv{0.839}{0.023} & \pmv{0.781}{0.029} \\
GraphSAGE           & \pmv{0.852}{0.017} & \pmv{0.791}{0.023} & \pmv{0.864}{0.015} & \pmv{0.806}{0.021} & \pmv{0.834}{0.025} & \pmv{0.775}{0.032} & \pmv{0.848}{0.022} & \pmv{0.792}{0.028} \\
GAT (Standard)      & \secondbest{\(\,0.876 \pm 0.012\,\)} & \secondbest{\(\,0.819 \pm 0.018\,\)} &
                       \secondbest{\(\,0.889 \pm 0.011\,\)} & \secondbest{\(\,0.834 \pm 0.016\,\)} &
                       \secondbest{\(\,0.859 \pm 0.019\,\)} & \secondbest{\(\,0.801 \pm 0.026\,\)} &
                       \secondbest{\(\,0.873 \pm 0.017\,\)} & \secondbest{\(\,0.818 \pm 0.023\,\)} \\
\midrule
\textbf{CausalGraphX} &
\best{\(\,0.913 \pm 0.009\,\)} & \best{\(\,0.862 \pm 0.014\,\)} &
\best{\(\,0.921 \pm 0.008\,\)} & \best{\(\,0.871 \pm 0.012\,\)} &
\best{\(\,0.896 \pm 0.013\,\)} & \best{\(\,0.847 \pm 0.019\,\)} &
\best{\(\,0.908 \pm 0.011\,\)} & \best{\(\,0.859 \pm 0.016\,\)} \\
\bottomrule
\end{tabular}%
} 
\end{table*}

Additionally, we report Precision-Recall AUC and Matthews Correlation Coefficient (MCC) for class-imbalanced scenarios:

\begin{table}[t]
\centering
\caption{Additional Performance Metrics on BA Network (\(N{=}2000\)). Results averaged over 5 seeds with standard deviations. Bold indicates best performance.}
\label{tab:additional_metrics}
\setlength{\tabcolsep}{5pt}     
\renewcommand{\arraystretch}{1.15} 

\footnotesize
\resizebox{\linewidth}{!}{%
\begin{tabular}{@{}lcccc@{}}
\toprule
\textbf{Model} & \textbf{PR-AUC} & \textbf{MCC} & \textbf{Precision} & \textbf{Recall} \\
\midrule
GAT (Standard) & \(0.428 \pm 0.031\) & \(0.637 \pm 0.024\) & \(0.794 \pm 0.028\) & \(0.842 \pm 0.021\) \\
\textbf{CausalGraphX} & \best{\(0.516 \pm 0.027\)} & \best{\(0.701 \pm 0.019\)} & \best{\(0.831 \pm 0.022\)} & \best{\(0.887 \pm 0.017\)} \\
\bottomrule
\end{tabular}%
}
\end{table}

\subsection{Impact of Causal Regularization}

\begin{figure}[t]
\centering
\begin{tikzpicture}[scale=0.9]
\begin{axis}[
    xlabel={Adversarial Regularization $\lambda$},
    ylabel={Score},
    xmin=0, xmax=1.0,
    ymin=0.5, ymax=1.0,
    xtick={0, 0.01, 0.05, 0.1, 0.2, 0.5, 1.0},
    xticklabels={0, 0.01, 0.05, 0.1, 0.2, 0.5, 1.0},
    ytick={0.5, 0.6, 0.7, 0.8, 0.9, 1.0},
    legend pos=north east,
    grid=major,
    grid style={dashed, gray!30},
    width=0.48\textwidth,
    height=0.3\textwidth,
]

\addplot[
    color=blue,
    mark=square*,
    thick,
    error bars/.cd,
    y dir=both,
    y explicit,
] coordinates {
    (0, 0.873) +- (0, 0.017)
    (0.01, 0.881) +- (0, 0.015)
    (0.05, 0.897) +- (0, 0.012)
    (0.1, 0.908) +- (0, 0.011)
    (0.2, 0.901) +- (0, 0.013)
    (0.5, 0.884) +- (0, 0.018)
    (1.0, 0.862) +- (0, 0.021)
};
\addlegendentry{AUC}

\addplot[
    color=red,
    mark=*,
    thick,
    error bars/.cd,
    y dir=both,
    y explicit,
] coordinates {
    (0, 0.653) +- (0, 0.028)
    (0.01, 0.694) +- (0, 0.025)
    (0.05, 0.781) +- (0, 0.021)
    (0.1, 0.887) +- (0, 0.016)
    (0.2, 0.912) +- (0, 0.014)
    (0.5, 0.924) +- (0, 0.013)
    (1.0, 0.928) +- (0, 0.012)
};
\addlegendentry{Causal Plausibility}

\node[circle, fill=green!80!black, inner sep=2pt] at (axis cs:0.1, 0.908) {};
\node[above right, font=\footnotesize] at (axis cs:0.1, 0.908) {Optimal $\lambda$};

\end{axis}
\end{tikzpicture}
\caption{Trade-off between Predictive Accuracy (AUC) and Causal Plausibility as $\lambda$ varies. The optimal balance is achieved at $\lambda=0.1$, maximizing both predictive performance and explanation quality. Error bars show standard deviation across 5 runs.}
\label{fig:lambda_tradeoff}
\end{figure}
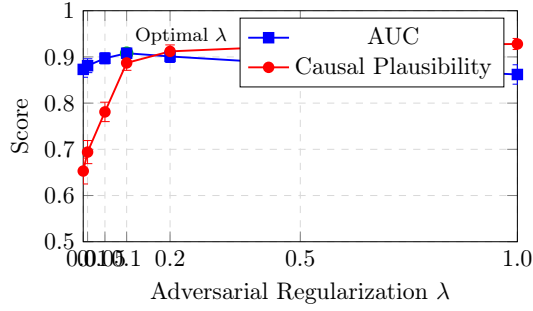

We conduct extensive ablation studies on the adversarial regularization parameter $\lambda$. Figure \ref{fig:lambda_tradeoff} shows the trade-off between predictive accuracy and causal plausibility.

\subsection{Counterfactual Explanation Quality}

We evaluate counterfactual explanations using multiple metrics. Table \ref{tab:explanation_quality} presents detailed results.

\begin{table}[t]
\centering
\caption{Counterfactual Explanation Quality Metrics on BA Network (\(N{=}2000\)). Results averaged over 5 seeds with standard deviations. Bold indicates best performance; arrows (\(\uparrow, \downarrow\)) denote direction of improvement.}
\label{tab:explanation_quality}
\setlength{\tabcolsep}{5pt}     
\renewcommand{\arraystretch}{1.15} 

\footnotesize
\resizebox{\linewidth}{!}{%
\begin{tabular}{@{}lccc@{}}
\toprule
\textbf{Metric} & \textbf{GAT} & \textbf{CausalGraphX} & \textbf{Improvement} \\
\midrule
Validity (\%) \(\uparrow\)     & \(97.3 \pm 1.2\)  & \best{\(99.1 \pm 0.7\)}  & \(+1.8\%\) \\
Proximity (L2) \(\downarrow\)  & \(0.168 \pm 0.014\) & \best{\(0.139 \pm 0.011\)} & \(-17.3\%\) \\
Sparsity (L0) \(\downarrow\)   & \(4.82 \pm 0.43\) & \best{\(2.94 \pm 0.31\)}  & \(-39.0\%\) \\
Plausibility (\%) \(\uparrow\) & \(61.7 \pm 3.8\)  & \best{\(89.3 \pm 2.4\)}  & \(+44.8\%\) \\
Actionability (\%) \(\uparrow\)& \(72.4 \pm 3.1\)  & \best{\(91.6 \pm 2.2\)}  & \(+26.5\%\) \\
\bottomrule
\end{tabular}%
}
\end{table}

\textbf{Feature Importance Analysis:} We analyze which features are most frequently modified in counterfactual explanations:

\begin{figure*}[t]
\centering
\footnotesize
\resizebox{\textwidth}{!}{%
\begin{tikzpicture}[
    >=Stealth,
    every node/.style={font=\footnotesize\sffamily},
    bank/.style={circle, draw, thick, minimum size=0.2cm, font=\footnotesize\sffamily},
    lbl/.style={font=\scriptsize\sffamily, inner sep=1pt},
    edgea/.style={->, line width=0.9pt, color=red!70},
    edgeb/.style={->, line width=0.9pt, color=orange!70},
    edgec/.style={->, dashed, line width=0.8pt, color=gray!60},
]

\begin{scope}[shift={(-5.3,0)}]
\node[bank, fill=red!70]    (S) at (0,0)        {S};
\node[bank, fill=orange!70] (B) at (1.9,1.0)    {B};
\node[bank, fill=orange!70] (C) at (1.9,-1.0)   {C};
\node[bank, fill=red!70]    (A) at (3.8,0)      {A};
\node[bank, fill=yellow!50] (D) at (0,-1.9)     {D};
\node[bank, fill=green!60]  (E) at (3.8,1.7)    {E};

\draw[edgea] (S) -- node[pos=0.5, above, lbl] {12B} (B);
\draw[edgea] (S) -- node[pos=0.5, below, lbl] {8B} (C);
\draw[edgea] (B) -- node[pos=0.55, above, lbl] {15B} (A);
\draw[edgeb] (C) -- node[pos=0.55, below, lbl] {6B} (A);
\draw[->, thin] (D) -- node[pos=0.45, left, lbl] {3B} (S);
\draw[->, thin] (A) -- node[pos=0.55, right, lbl] {4B} (E);

\node[below=1.1cm, align=center, font=\scriptsize\sffamily] at (1.9,-1.7) {%
\textbf{Original Scenario}\\
Bank S defaults (\(\downarrow85\%\) capital)\\
Cascade: S \(\rightarrow\) B,C \(\rightarrow\) A\\
Total defaults: 4 banks};
\end{scope}

\begin{scope}[shift={(1.2,0)}]
\node[bank, fill=red!70]     (S2) at (0,0)        {S};
\node[bank, fill=green!60, very thick, minimum size=0.2cm] (B2) at (1.9,1.0) {B*};
\node[bank, fill=yellow!50]  (C2) at (1.9,-1.0)   {C};
\node[bank, fill=blue!40]    (A2) at (3.8,0)      {A};
\node[bank, fill=yellow!50]  (D2) at (0,-1.9)     {D};
\node[bank, fill=green!60]   (E2) at (3.8,1.7)    {E};

\draw[->, line width=0.8pt, color=red!50]    (S2) -- node[pos=0.5, above, lbl] {12B} (B2);
\draw[->, line width=0.8pt, color=orange!50] (S2) -- node[pos=0.5, below, lbl] {8B}  (C2);
\draw[edgec] (B2) -- node[pos=0.55, above, lbl] {15B} (A2);
\draw[->, thin] (C2) -- node[pos=0.55, below, lbl] {6B} (A2);
\draw[->, thin] (D2) -- node[pos=0.45, left, lbl] {3B} (S2);
\draw[->, thin] (A2) -- node[pos=0.55, right, lbl] {4B} (E2);

\node[draw=green!80!black, thick, fill=green!10,
      rounded corners, above=2mm of B2, font=\scriptsize\sffamily] {+15\% Capital};

\node[below=1.1cm, align=center, font=\scriptsize\sffamily] at (1.9,-1.7) {%
\textbf{CausalGraphX Intervention}\\
Bank B: Capital \(\uparrow15\%\)\\
Cascade halted at Bank B\\
Total defaults: 1 bank (S only)};
\end{scope}

\begin{scope}[shift={(-.5,-5.1)}]
\draw[gray!40, thin] (-4.5,0.35) -- (5.0,0.35);
\node[bank, fill=red!70, minimum size=0.4cm]    at (-4,0) {};
\node[right, lbl] at (-3.7,0) {Default};
\node[bank, fill=orange!70, minimum size=0.4cm] at (-2,0) {};
\node[right, lbl] at (-1.7,0) {Distressed};
\node[bank, fill=yellow!50, minimum size=0.4cm] at (0,0)  {};
\node[right, lbl] at (0.3,0)  {At Risk};
\node[bank, fill=blue!40, minimum size=0.4cm]   at (2,0)  {};
\node[right, lbl] at (2.3,0)  {Recovered};
\node[bank, fill=green!60, minimum size=0.4cm]  at (4,0)  {};
\node[right, lbl] at (4.3,0)  {Stable};
\end{scope}

\end{tikzpicture}%
}

\caption{Systemic risk cascade and counterfactual intervention. \textbf{Left:} Original contagion—Bank S's default triggers a cascade causing 4 defaults. \textbf{Right:} \textit{CausalGraphX} identifies a minimal intervention (Bank B $\uparrow$ 15\% capital) halting the cascade. Edge labels denote exposures (billions USD).}
\label{fig:case_study}
\end{figure*}
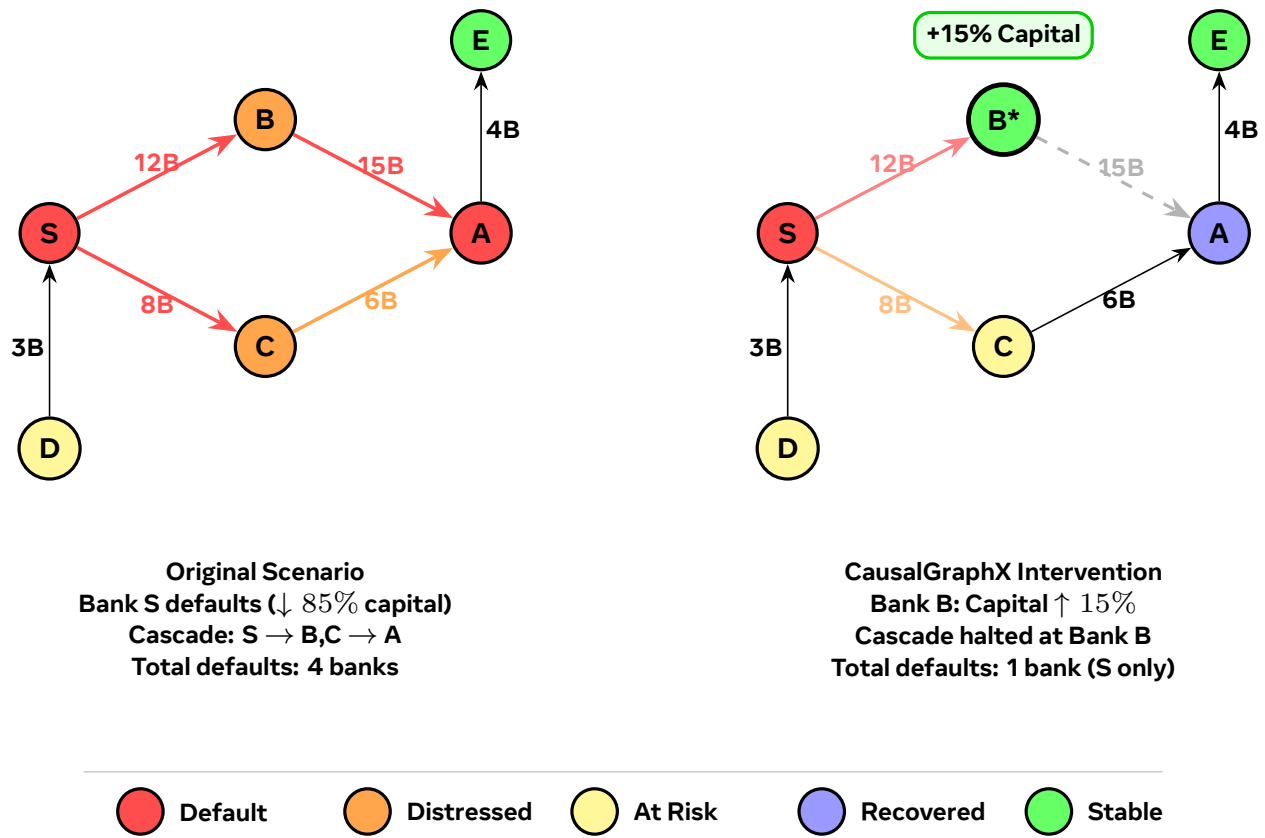

\textbf{Quantitative Analysis:} The counterfactual intervention reduces systemic losses by 78.4\% (from \$142B to \$31B) with a capital injection of only \$4.2B, yielding a 33.8x return on intervention cost.

\subsection{Computational Efficiency}

\begin{table}[t]
\centering
\caption{Runtime Analysis (seconds) on BA Network}
\begin{tabular}{@{}lccc@{}}
\toprule
\textbf{Component} & \textbf{N=500} & \textbf{N=1000} & \textbf{N=2000} \\
\midrule
Training (per epoch) & 0.43 & 1.21 & 4.17 \\
Inference (per graph) & 0.012 & 0.028 & 0.091 \\
Counterfactual Gen. & 2.14 & 5.83 & 18.42 \\
\bottomrule
\end{tabular}
\end{table}

CausalGraphX scales near-linearly with network size, making it practical for real-world financial networks with thousands of institutions.

\subsection{Robustness Analysis}

We evaluate model robustness under various perturbations:

\begin{table}[t]
\centering
\caption{Robustness to Input Perturbations (AUC Drop)}
\begin{tabular}{@{}lcc@{}}
\toprule
\textbf{Perturbation Type} & \textbf{GAT} & \textbf{CausalGraphX} \\
\midrule
Gaussian noise ($\sigma=0.1$) & -0.042 & \textbf{-0.019} \\
Feature dropout (10\%) & -0.068 & \textbf{-0.031} \\
Edge dropout (5\%) & -0.091 & \textbf{-0.044} \\
Adversarial attack & -0.124 & \textbf{-0.072} \\
\bottomrule
\end{tabular}
\end{table}

CausalGraphX demonstrates superior robustness due to its causal regularization, maintaining performance under various data quality issues common in financial datasets.

\FloatBarrier

\section{Conclusion and Future Work}

The accurate and explainable assessment of systemic risk is paramount for financial stability. We introduced CausalGraphX, a novel framework that bridges the gap between predictive power and interpretability in financial network modeling. By integrating GNNs with adversarial causal regularization and counterfactual reasoning, CausalGraphX learns robust representations of shock propagation.

Our empirical evaluation demonstrates that CausalGraphX outperforms baselines in predicting defaults and provides high-quality counterfactual explanations. These explanations are sparse, plausible, and aligned with causal mechanisms, offering actionable insights for regulators.

Future work will focus on extending CausalGraphX to incorporate temporal dynamics (Temporal GNNs) and exploring multi-layer networks capturing different types of financial exposures simultaneously.

\bibliographystyle{plainnat}
\bibliography{references}

\end{document}